\documentclass[journal, compsoc]{IEEEtran}
\usepackage{etex}
\ifCLASSOPTIONcompsoc
\usepackage[nocompress]{cite}
\else
\usepackage{cite}
\fi
\ifCLASSINFOpdf
\else
\fi
\usepackage{latexsym}
\usepackage{amsmath, epstopdf}
\usepackage{amssymb}
\usepackage{commath}
\usepackage[none]{hyphenat}
\usepackage[linesnumbered,ruled,vlined]{algorithm2e}
\usepackage{algpseudocode}
\usepackage{graphicx, subfigure}
\usepackage{tikz}
\usepackage{multirow}
\usepackage[counterclockwise]{rotating}
\usetikzlibrary{decorations.pathreplacing,calc}

\usepackage{array, booktabs, caption}
\usepackage{makecell}
\usepackage{adjustbox}
\usepackage{lineno,hyperref}
\usepackage[misc]{ifsym}
\usepackage{balance}
\usepackage{gensymb}

\def\etal{\emph{et~al.~}}
\begin{document}
\title{Localization of Unmanned Aerial Vehicles in Corridor Environments using Deep Learning}
\author{Ram~Prasad~Padhy,
	Shahzad~Ahmad,
	Sachin~Verma,
	Pankaj~Kumar~Sa,~\IEEEmembership{Member,~IEEE,}
	Sambit~Bakshi,~\IEEEmembership{Member,~IEEE}
	\IEEEcompsocitemizethanks{\IEEEcompsocthanksitem R. P. Padhy, S. Ahmad, S. Verma, P. K. Sa and S. Bakshi are with the Department of Computer Science and Engineering, National Institute of Technology, Rourkela 769008, India. 
		(E-mail: ramprasad.nitr@gmail.com, shahzadnit@gmail.com, jsrsachinverma999@gmail.com , sambitbaksi@gmail.com/bakshisamsbit@ieee.org, pankajksa@nitrkl.ac.in)
	}
}

\IEEEtitleabstractindextext{%
	\begin{abstract}
		Vision-based pose estimation of Unmanned Aerial Vehicles (UAV) in unknown environments is a rapidly growing research area in the field of robot vision. The task becomes more complex when the only available sensor is a static single camera (monocular vision). In this regard, we propose a monocular vision assisted localization algorithm, that will help a UAV to navigate safely in indoor corridor environments. Always, the aim is to navigate the UAV through a corridor in the forward direction by keeping it at the center with no orientation either to the left or right side. The algorithm makes use of the RGB image, captured from the UAV front camera, and passes it through a trained deep neural network (DNN) to predict the position of the UAV as either on the left or center or right side of the corridor. Depending upon the divergence of the UAV with respect to the central bisector line (CBL) of the corridor, a suitable command is generated to bring the UAV to the center. When the UAV is at the center of the corridor, a new image is passed through another trained DNN to predict the orientation of the UAV with respect to the CBL of the corridor. If the UAV is either left or right tilted, an appropriate command is generated to rectify the orientation. We also propose a new corridor dataset, named NITRCorrV1, which contains images as captured by the UAV front camera when the UAV is at all possible locations of a variety of corridors. An exhaustive set of experiments in different corridors reveal the efficacy of the proposed algorithm.
	\end{abstract}
	\begin{flushleft}
		
	\end{flushleft}
}
\maketitle	
\IEEEdisplaynontitleabstractindextext
\IEEEpeerreviewmaketitle
\IEEEraisesectionheading{\section{Introduction}\label{intro}}
\IEEEPARstart{I}n recent years, Unmanned Aerial Vehicles (UAV) equipped with a single or multiple cameras have gained huge popularity in computer vision research domain due to their unique ability to present a scene from an aerial perspective. Nowadays, UAVs are predominantly used in many fields, such as aerial surveillance, military applications, inspection of disaster-hit areas or earthquake affected buildings, precision agriculture, movie making, bridge inspection, and many more. However, an accurate localization algorithm is essential in carrying out these tasks autonomously with absolute aplomb. In most of the outdoor environments, Global Positioning System (GPS) helps in obtaining the pose of the UAV and thus helps in autonomous navigation~\cite{gryte2017robust}. On the contrary, unreliability or complete unavailability of GPS in most of the indoor environments makes the task of pose estimation very complex.
\par Several solutions, such as Simultaneous Localization and Mapping (SLAM)~\cite{chen2013designing,engel2012camera,engel2014scale}, Parallel Tracking and Mapping (PTAM)~\cite{sa2013monocular}, stereo vision \textit{etc.} have been proposed for indoor autonomous navigation of robots. In SLAM and PTAM, laser range finders, RGB-D sensors or a single camera are generally used to construct the 3D model of the surrounding environment. At any instant, the position of the robot on the map can be easily inferred for autonomous navigation. However, these tracking based systems require real-time heavy computation for constructing the map making them more suitable for ground robots as compared to aerial systems. In the case of stereo vision, depth is estimated by constructing the disparity map from stereo images~\cite{achtelik2009stereo, mcguire2017efficient, qin2016stereo}. However, these systems fail spectacularly in textureless environments, such as walls where it is very difficult to obtain features for disparity calculation. Hence, stereo based systems cannot be implemented in the case of corridor environments. Moreover, most of the available UAVs in the market are equipped with a single camera, and hence stereo vision solution is not practical in general.
\par In this paper, we present a monocular vision assisted method for localization and subsequently safe navigation of a UAV in corridor environments. The UAV in motion is required to avoid side-wise collision with the side-walls as well as should cease motion to avoid collision with the front wall, present at the end of the corridor. Our navigation algorithm takes input from the UAV front camera, which is a static pinhole camera. To achieve our goal, the UAV is required to localize itself in the unknown corridor. Our aim is to keep the UAV at the center of the corridor with no orientation either to the left or right. Once the algorithm safely localizes the UAV inside the corridor, it can navigate in the forward direction with minimal chance of collision. However, different external (e.g. wind) and internal hindrances (e.g. rotor turbulence) may deviate the direction during UAV flight. Hence, the algorithm continuously monitors the pose of the UAV during its flight and generates the necessary command in case there is any deviation from the actual path.
\par The advantage of this method over the existing PTAM and SLAM based systems is that it does not require real-time 3D construction of the environment, thus computation is very low. Also, the proposed localization algorithm can act as an additive to tracking based systems and significantly lower the computation time. Hence, the proposed algorithm is more suitable for UAV systems, where the reaction time is very low as compared to ground robots. Furthermore, unlike the stereo-based systems, this method is not dependent on feature extraction, hence can very well work in texture-less environments, such as corridors.
\par The principal contributions of this paper are as follows:\\
\textbf{(a)} We propose a model, which uses two similar deep neural networks (DNN) to localize the UAV in unknown indoor corridor environments. The first neural network is responsible for predicting the deviation of the UAV in terms of translation from the center of the corridor, whereas the second neural network predicts the orientation of the UAV, when the UAV is approximately at the center.\\
\textbf{(b)} The algorithm generates necessary control commands to keep the UAV along the central bisector line of the corridor and continuously monitors the position to rectify any deviation.\\
\textbf{(c)} We propose a new corridor dataset, NITRCorrV1~\cite{padhy2019dataset}, which contains images as captured from the UAV front camera from different positions of a number of corridors, having varying dimensions and intensity exposure. As per the best of our knowledge, we are the first ones to propose such a dataset, which contains the deviation of the UAV in terms of translational and rotational parameters.\\
\textbf{(d)} We show the efficacy of the proposed algorithm in real-time UAV navigation flights inside indoor corridors. 
\par Rest of the paper is organized as follows. Section~\ref{sec-literature} presents prior research in the field of UAV navigation. Section~\ref{sec-overview} provides an overview of the proposed methodology. Section~\ref{data-create} presents the detailed process of our custom dataset creation followed by a thorough elaboration on the proposed DNN based autonomous navigation algorithm in Section~\ref{sec-dnn}. Section~\ref{sec-expt} provides the experimental results of our proposed model along with the results on real-time UAV navigation experiments. Finally, the concluding remaks are given in Section~\ref{sec-conclusion}.                  
\section{Related Work}\label{sec-literature}
The research work in the field of UAV navigation has progressed quite significantly in the recent past. The prior work in the field of autonomous UAV navigation can be categorized into two main fields depending on the type of sensors used; \textbf{(a)} Navigation using proximity sensors, and \textbf{(b)} Navigation using vision sensors\\
\textbf{Navigation using Proximity Sensors: }Proximity sensors, such as LIDARs, ultrasonic sensors \textit{etc.} usually sense the obstacle in the vicinity with the help of infrared rays or electromagnetic radiations~\cite{grzonka2008autonomous,bachrach2011range,gageik2015obstacle}. Cruz \etal used the proximity sensors installed on a UAV to model the obstacle locations as well as the goal point in the environment~\cite{cruz2012obstacle}. A SLAM system on Android platform is developed by Chen \etal with the help of readings from the Inertial Measurement Unit (IMU) and ultrasonic sensors~\cite{chen2013designing}. A laser-guided system is proposed by Stubblebine \etal, which also uses LIDAR data to map the surrounding environment~\cite{stubblebine2015laser}. Roberts \etal developed a UAV navigation model, which employed a Kalman filter over the data from ultrasonic sensors to estimate the best path at each location on the basis of observation density, which is estimated from a predefined map of the environment~\cite{roberts2007quadrotor}. Although researchers have used proximity sensors to sense obstacles, these are not best suited for UAV navigation. High weight and difficulty in installation on a UAV make them infeasible in most of the cases of autonomous UAV navigation.\\
\textbf{Navigation using Vision Sensors:} Cameras are lightweight and low power consuming sensors, and hence is the ideal solution to detect obstacles in case of autonomous UAV navigation. In stereo vision (two cameras), triangulation geometry is used to obtain the disparity map from common feature points of the two camera views~\cite{achtelik2009stereo, mcguire2017efficient, qin2016stereo}. However, power constraint of the UAV system demands for a better solution, such as the use of a single camera (monocular vision). Monocular vision processing is more challenging as compared to stereo vision because of the absence of depth information. These systems usually employ visual odometry and optical flow to sense obstacles in unknown environments~\cite{martinez2015towards,mori2013first,sa2013monocular,saha2014real,wang2015obstacle,weiss2012real}. He \etal used the IMU readings along with the data from a monocular camera to model the ego-motion parameters~\cite{he2006vision}. Sa \etal proposed a Kalman filter based PTAM algorithm using the inputs from a static single camera~\cite{sa2013monocular}. Few researchers have stressed upon the use of vanishing point from corridor images to navigate in indoor environments~\cite{lioulemes2014safety, padhy2018monocular}. Kim and Chen~\cite{kim2015deep}, and Padhy \etal~\cite{padhy2018deep} in their respective works have developed DNN based systems, that are trained on corridor datasets, to predict the next command for autonomous UAV navigation in indoor corridors. An Extended Kalman filter based monocular SLAM system was developed by Engel \etal to approximate the UAV position in the surrounding environment~\cite{engel2012camera,engel2014scale}. A learning-based UAV navigation algorithm based on optical flow was developed by Wang \etal~\cite{wang2015obstacle}. Bill \etal developed a system to detect obstacles in a few indoor scenarios, such as closed rooms. corridors and staircases~\cite{bills2011autonomous}. A dead reckoning and histogram voting scheme for UAV navigation was developed by the researchers from Czech Technical University~\cite{etienne1999dead}. Few papers have stressed upon the use of visual keypoints and their scale change with the movement of the UAV towards the obstacle to predict the distance from a moving UAV~\cite{saha2014real,mori2013first,chavez2009vision}.     
\par Our current work uses a DNN based model that takes inputs from the UAV front camera to measure the translational as well as rotational deviation of the UAV inside a corridor. These deviations are rectified throughout the process of UAV navigation to localize the UAV safely. Unlike other neural network based UAV navigation methods inside the corridor~\cite{kim2015deep, padhy2018deep}, that give predictions in terms of flight commands, our model predicts the approximate deviations of the UAV from a safe path. We also proposed a new corridor dataset to accomplish the task.
\begin{figure}[!htb]
	\centering
	\includegraphics[width=0.48\textwidth, height=0.20\textheight]{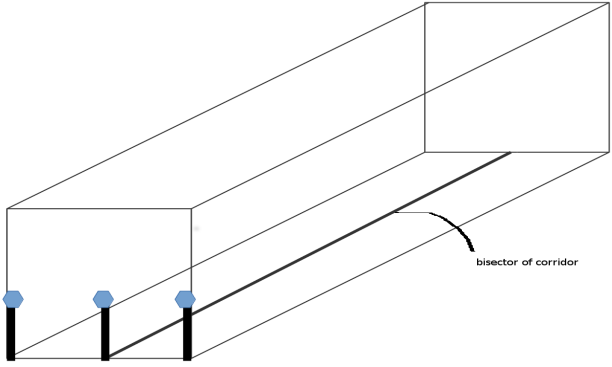}
	\caption{Central bisector line of a corridor.}
	\label{bisector}
\end{figure}
\section{Overview of the Solution Strategy}\label{sec-overview}
Our goal is to autonomously navigate a UAV in indoor corridors without any collision either with the side walls or with the front wall at the end of the corridor. The real-time data from the UAV front camera acts as input to our vision-based navigation algorithm. The algorithm first forwards one image frame through a trained regression network to measure the translational deviation of the UAV from an imaginary line, known as the central bisector line (CBL) of the corridor (Figure~\ref{bisector}). This line runs on the middle of the floor across the whole length of the corridor and is parallel to the side walls. It may be noted that there does not exist any such line in real-world corridor environments; rather it is an imaginary line, which is used as a reference to measure the deviation of the UAV from the center of the corridor. Depending on the deviation of the UAV with respect to the CBL, a suitable command is generated to bring the UAV to the center of the corridor. Again, another image frame is forwarded through a similar trained regression network to measure the orientation from the CBL. This variation in orientation, if any, is rectified by generating a suitable control command. Once, the UAV is aligned with the CBL, it navigates in the forward direction. The process of rectifying the translational and rotational deviation continues across the whole length of the corridor, as the UAV may be subjected to different external (e.g. wind) or internal (e.g. rotor turbulence) factors, which have the potential to deviate it from a straight line path.
\section{Dataset Creation}\label{data-create}
There exist many indoor datasets(~\cite{quattoni2009recognizing, huitl2012tumindoor, kim2015deep}) for the task of UAV navigation. However, none of them provides ground truth values in terms of deviation with respect to the CBL of the corridor, which according to us plays a significant role for localization of a UAV inside corridor environments. Hence, we provide our own custom dataset for this purpose.
\subsection{Image Capturing}
\begin{figure*}[!t]  
	\subfigure{\includegraphics[width=.33\textwidth]{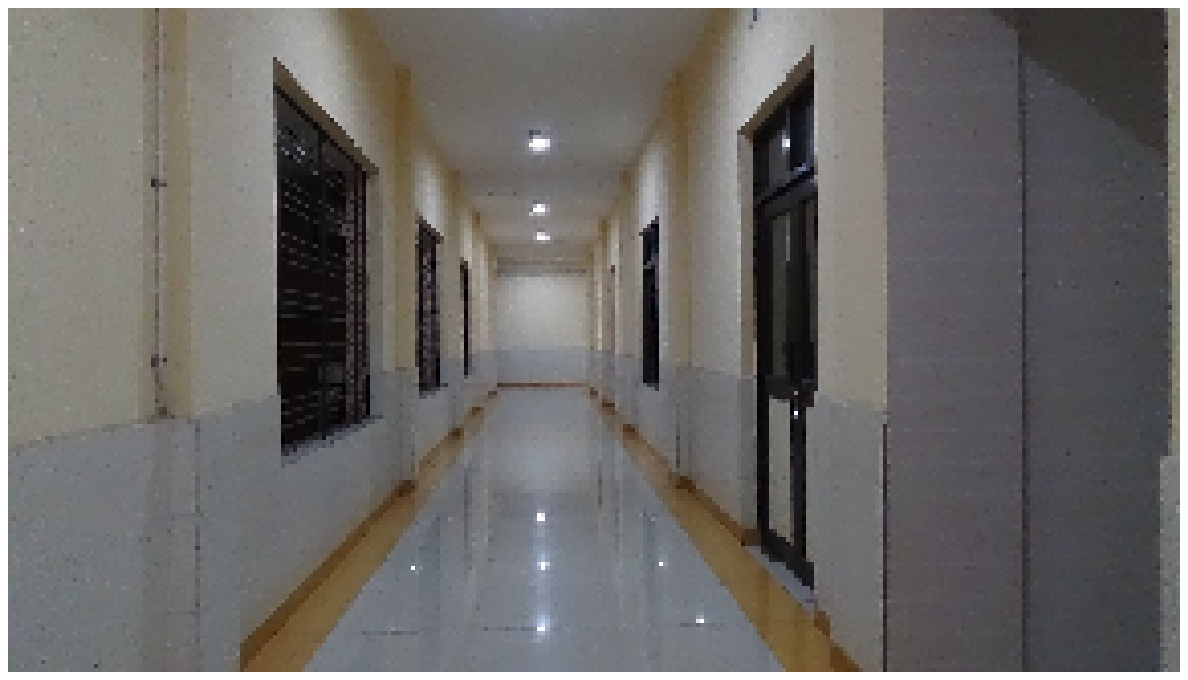}} 	
	\subfigure{\includegraphics[width=.33\textwidth]{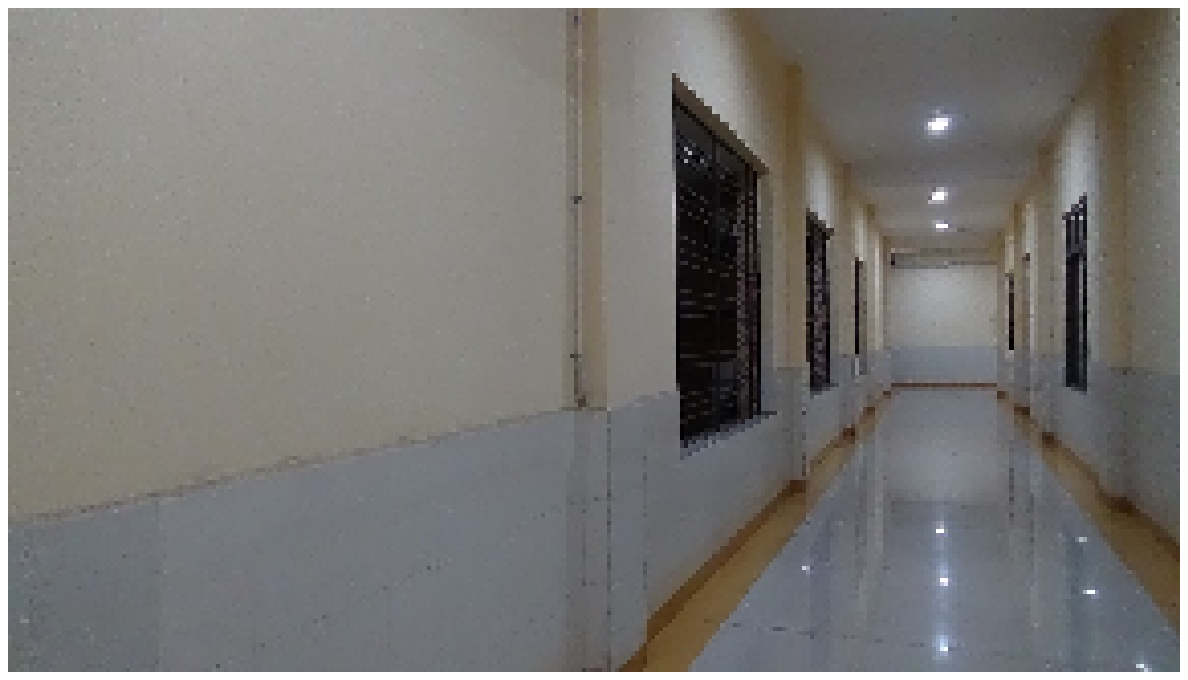}}
	\subfigure{\includegraphics[width=.33\textwidth]{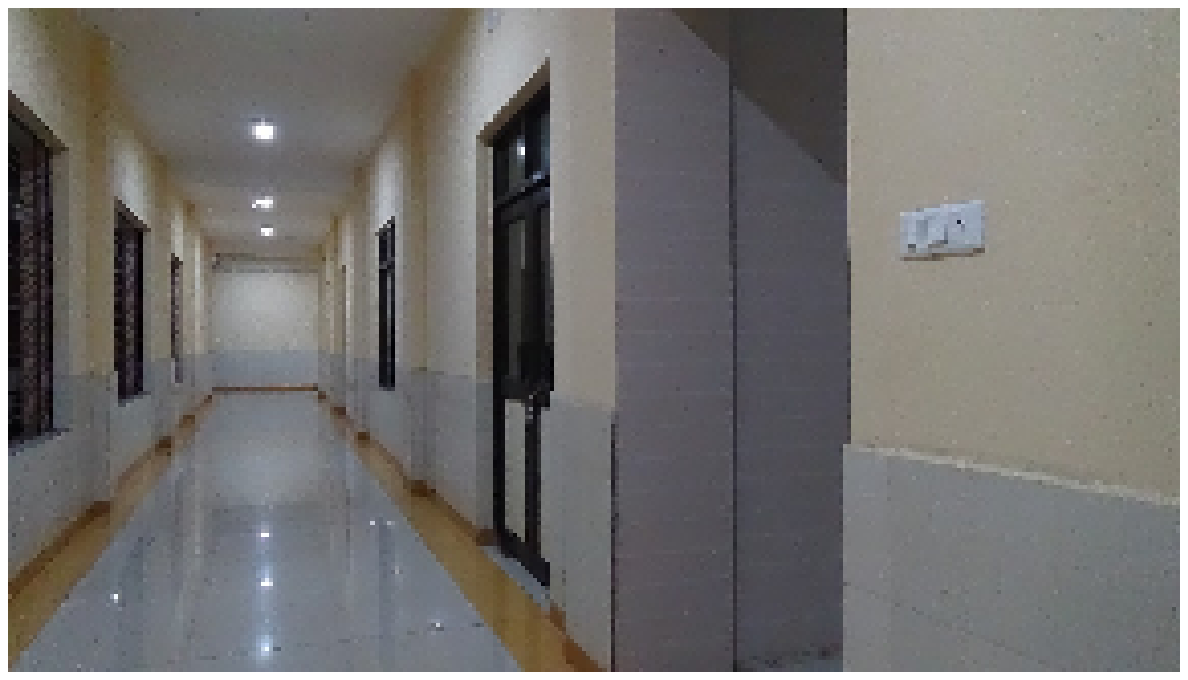}}\\
	\subfigure{\includegraphics[width=.33\textwidth]{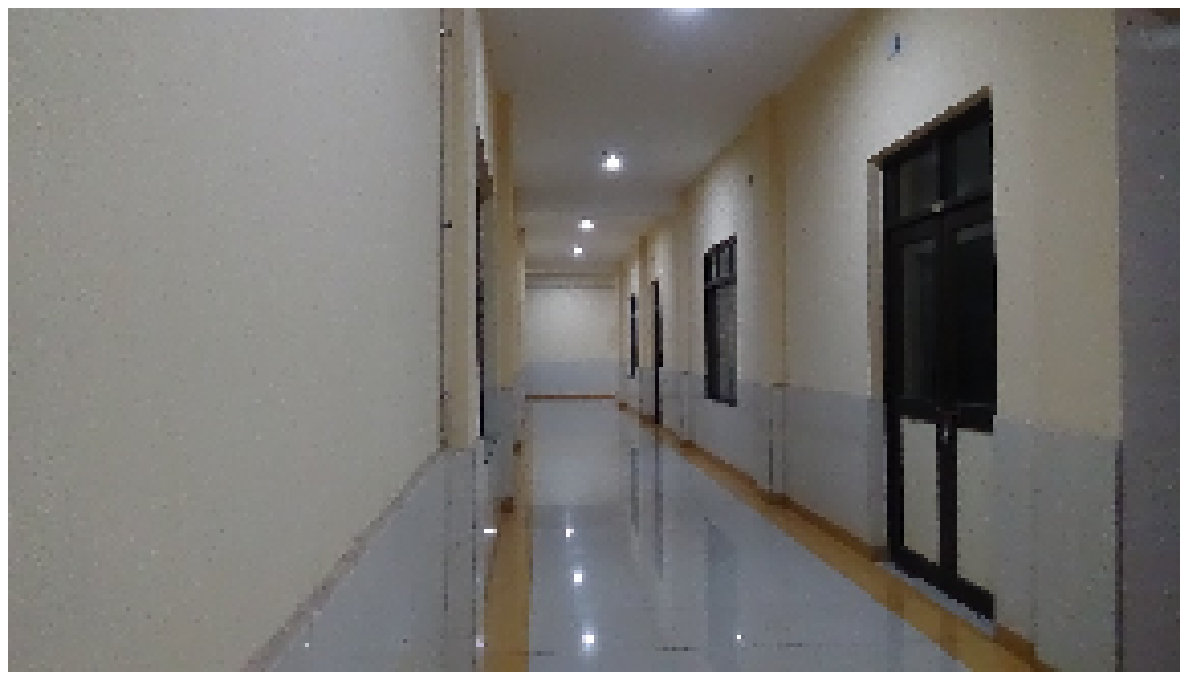}}
	\subfigure{\includegraphics[width=.33\textwidth]{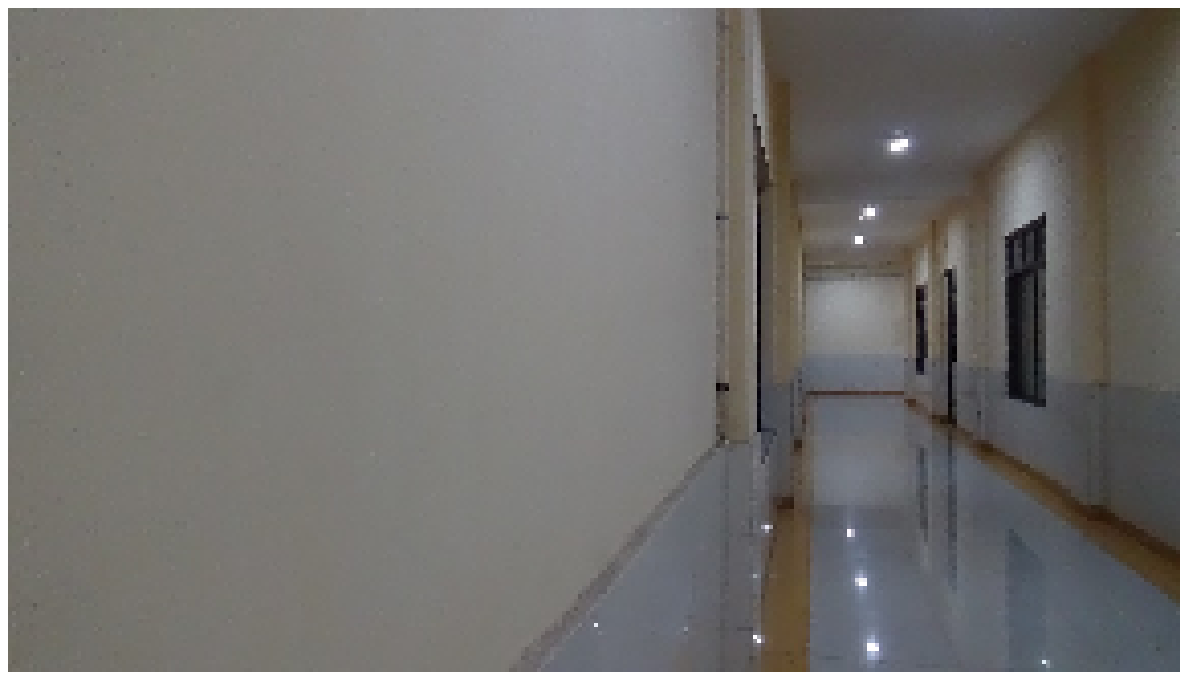}}
	\subfigure{\includegraphics[width=.33\textwidth]{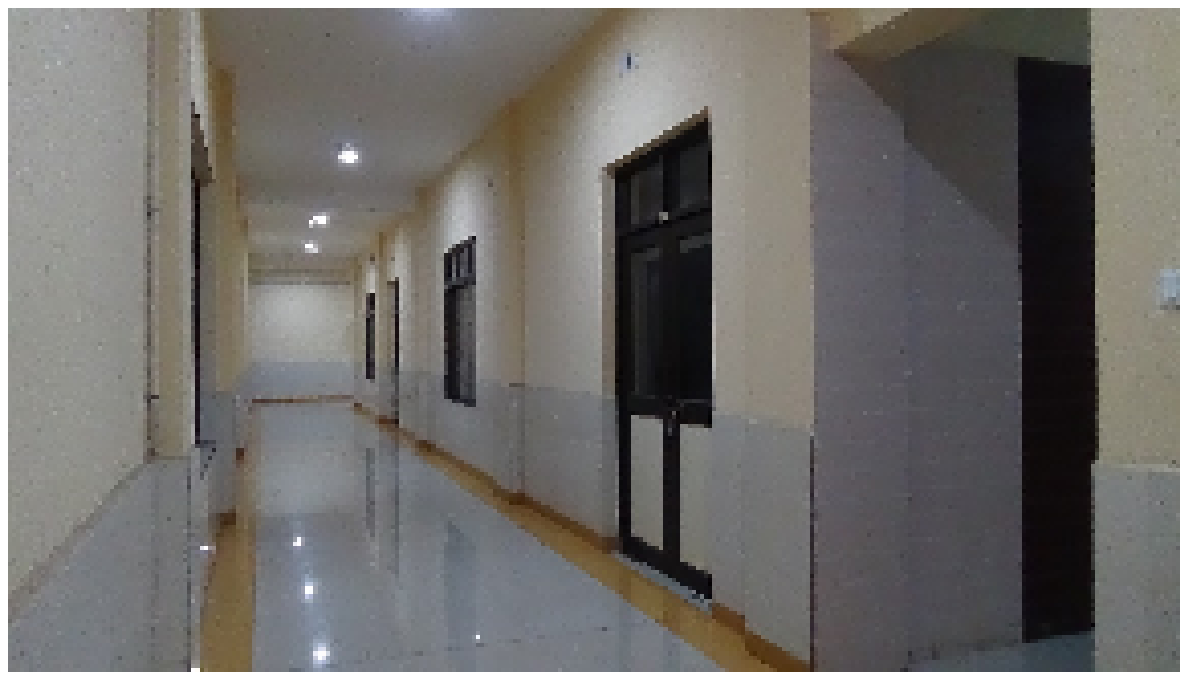}}\\
	\subfigure{\includegraphics[width=.33\textwidth]{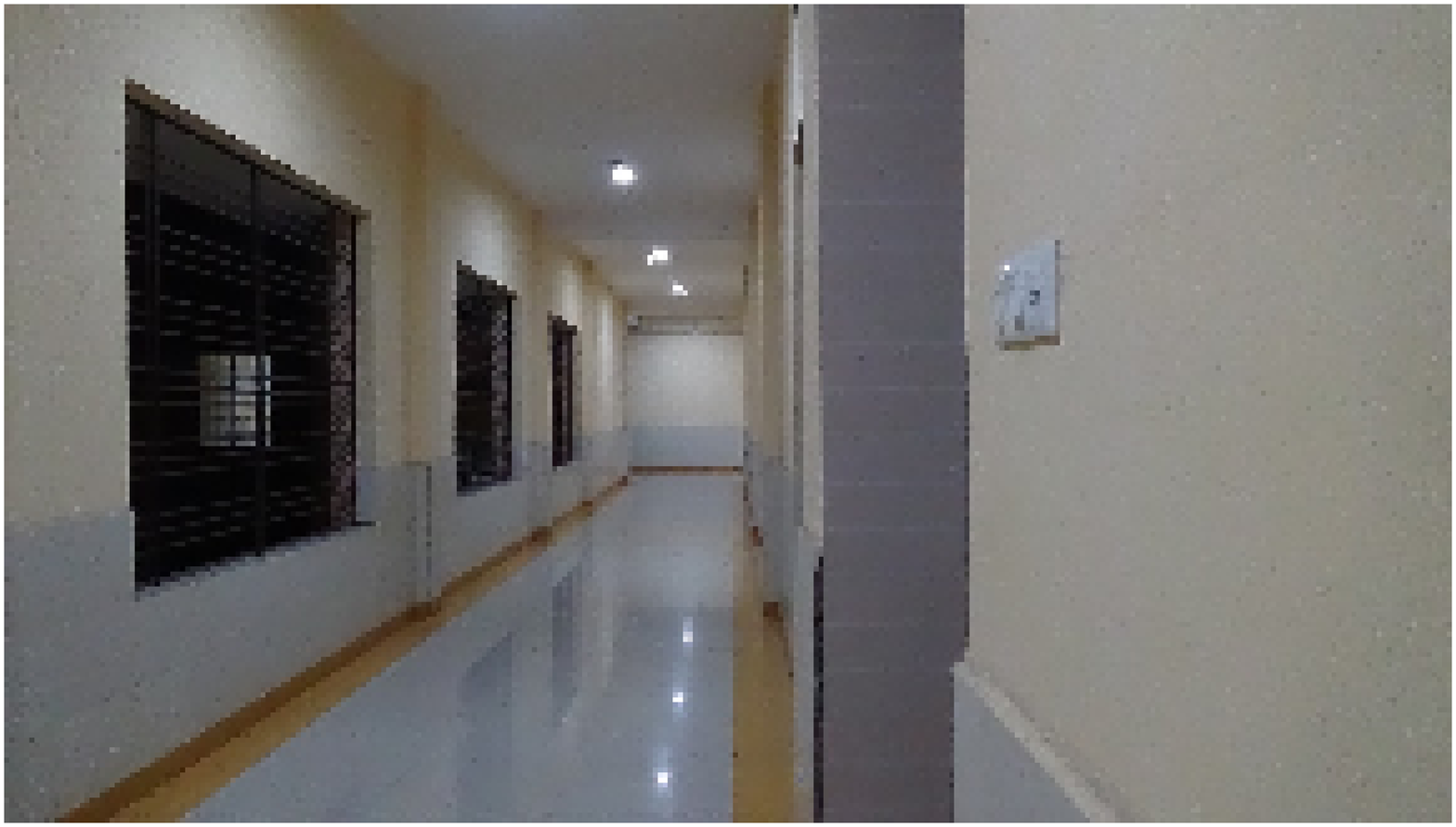}}
	\subfigure{\includegraphics[width=.33\textwidth]{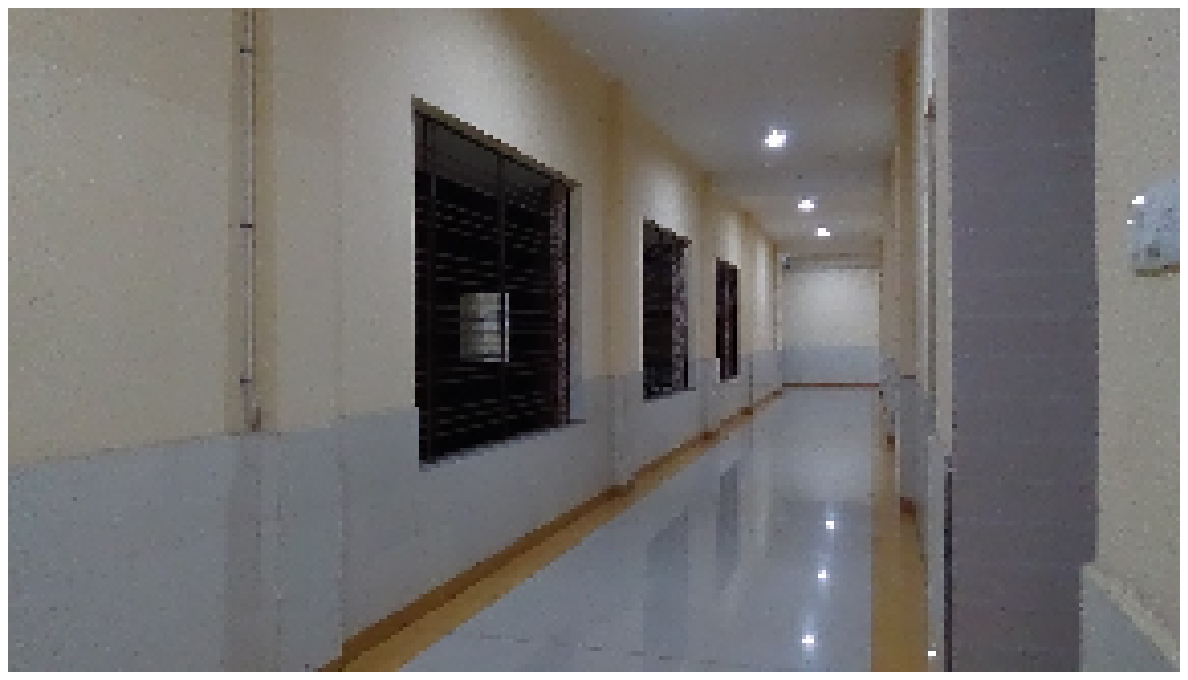}}
	\subfigure{\includegraphics[width=.33\textwidth]{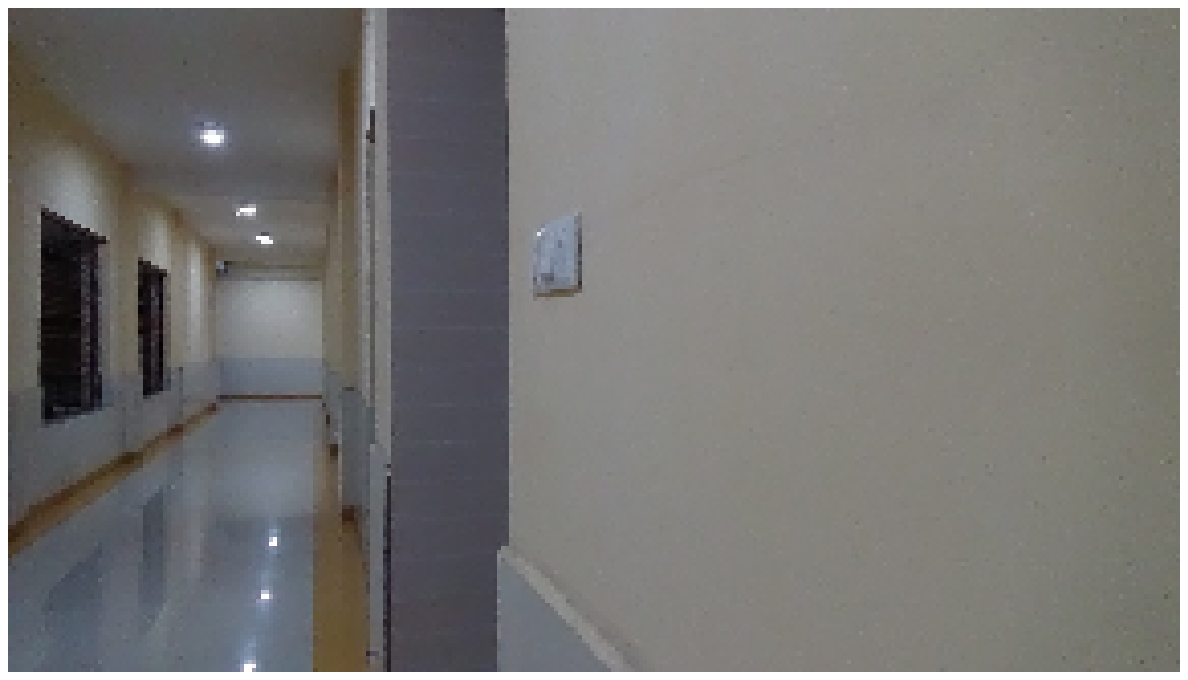}}
	\caption{Images as captured by the UAV front camera from $\mathrm{9}$ different possible alignments over a horizontal line perpendicular to the CBL. From top to bottom, the rows represent the images, when the UAV is at the center, extreme left and extreme right over the horizontal line, respectively. From left to right, the columns represent the orientations towards center, left and right, respectively.}
	\label{9-locations}
\end{figure*}

We captured images across $80$ different corridors of National Institute of Technology, Rourkela, India, out of which $59$ and $21$ corridors are used for training and testing purpose respectively. The images are captured by the UAV front camera at a resolution of $320 \times 180$. While preparing the dataset, it is made sure that the dataset covers all the possible positions across the whole length of the corridor. The height of the UAV is kept at approximately 1 meter. At a certain place inside the corridor, there different locations over a horizontal line, which is perpendicular to the CBL, are selected. These locations are mainly the center and two extreme sides of the corridor. At each location over the horizontal line, images are captured while the UAV is tilted in three different orientations: center, left, and right tilt. Hence, we will have $9$ different images corresponding to a particular horizontal line. Possible scenarios are shown in Figure~\ref{9-locations}. These horizontal lines are chosen across the whole length of the corridor at certain intervals. However, these images do not provide any information regarding the CBL. Hence, for each image, a similar image is captured by placing two markers on the CBL of the corridor. These images are known as bisector images. Markers are placed on the floor by measuring the middle point between the two side walls. One marker is placed at the extreme end of the corridor, and its position remains the same for all the images. The second marker is placed somewhere inside the corridor, such that it will be visible on the captured image. We alter the position of the second marker to make it visible throughout the process of image capturing. Although the actual image and its corresponding bisector image vary slightly in terms of pixel values, their geometry, which holds key information regarding the corridor structure, is preserved. It may be noted that the bisector images do not correspond to the final dataset. These are only used for obtaining the target values for the actual images.
\begin{figure*}[!htbp]   
	\subfigure[Actual image]{\includegraphics[width=.33\textwidth]{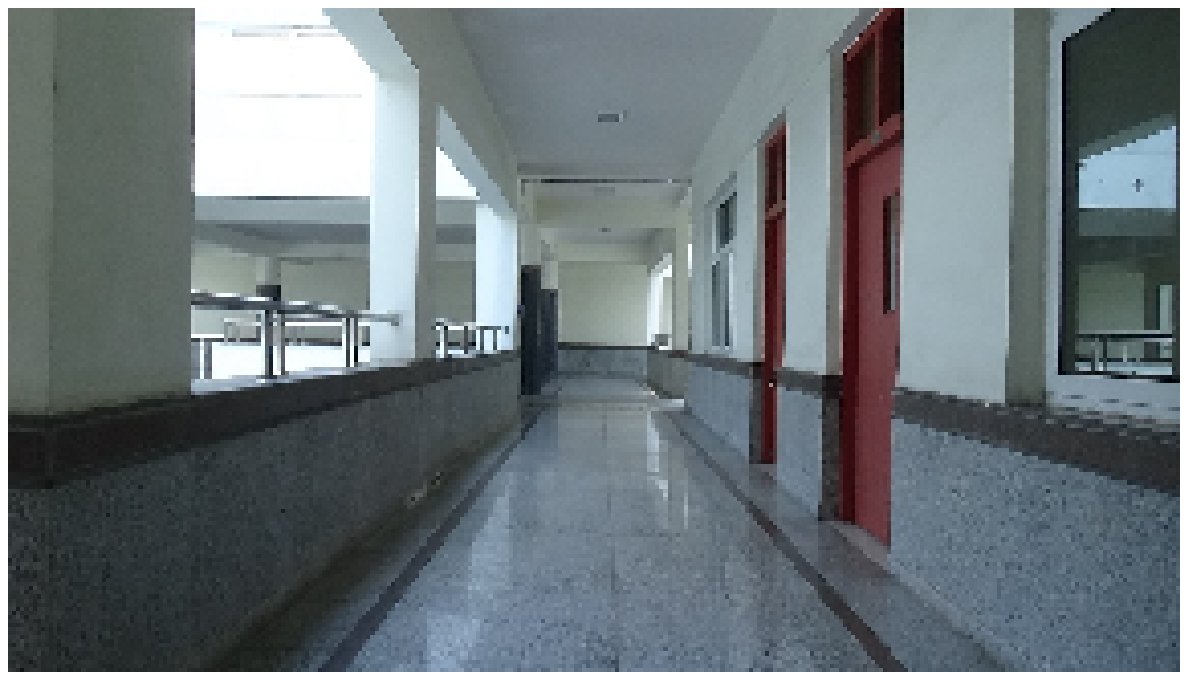}}
	\subfigure[Image with markers]{\includegraphics[width=.33\textwidth]{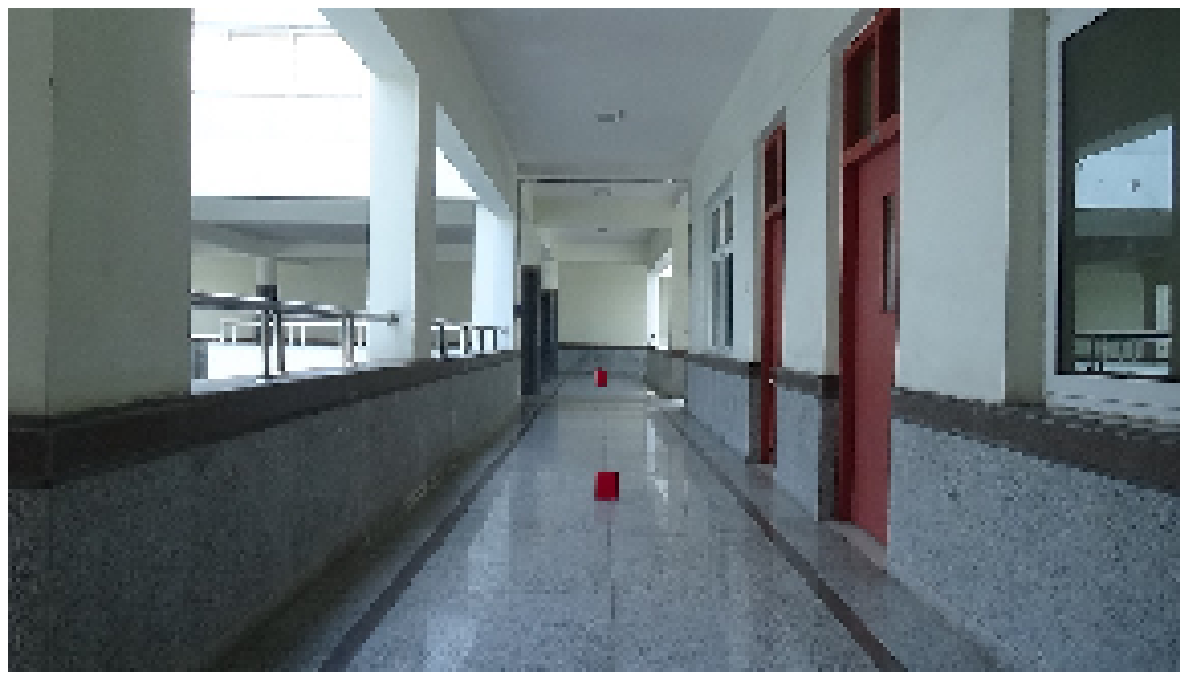}}
	\subfigure[CBL on image plane]{\includegraphics[width=.33\textwidth]{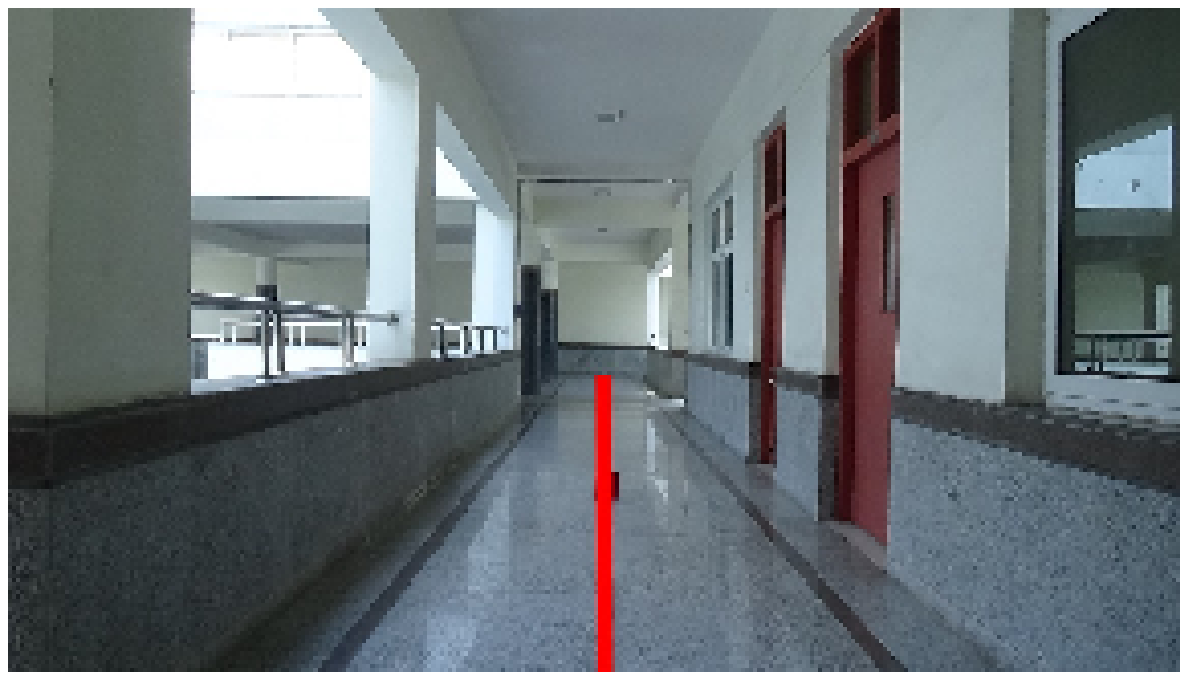}}
	\caption{Process of obtaining the CBL on the image plane using markers}
	\label{dataset}
\end{figure*}
\subsection{Labeled Data Generation}
\begin{figure*}[!htbp]   
	\subfigure[Left side of CBL]{\includegraphics[width=.33\textwidth]{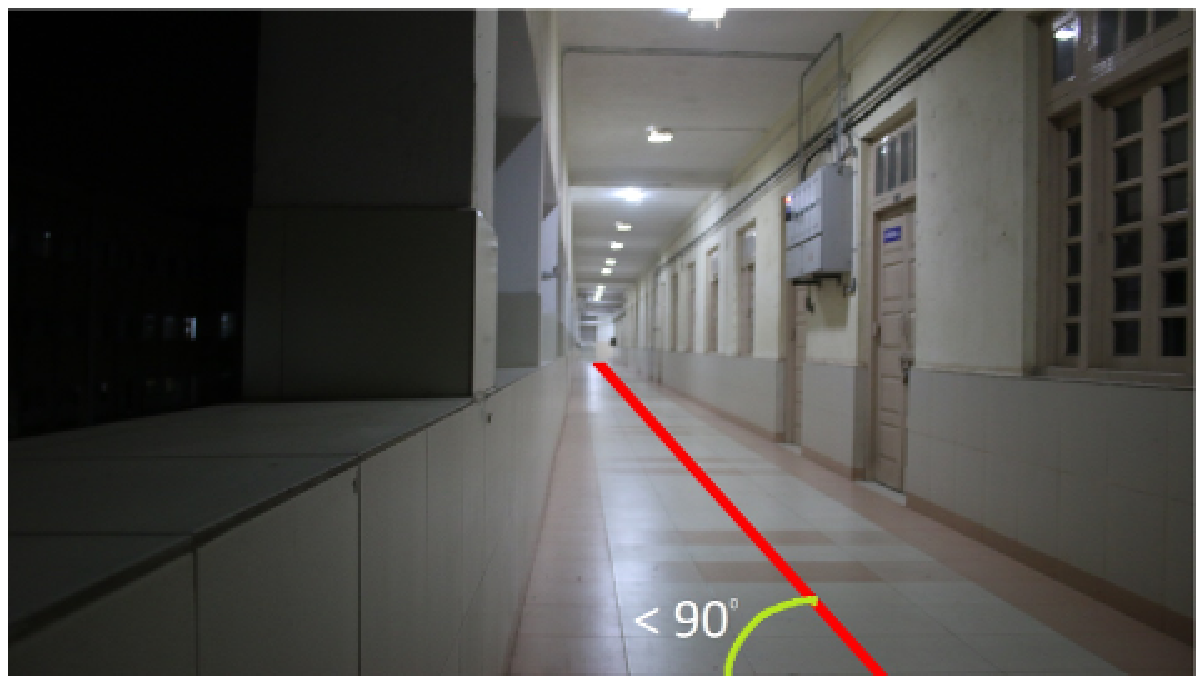}}
	\subfigure[Right side of CBL]{\includegraphics[width=.33\textwidth]{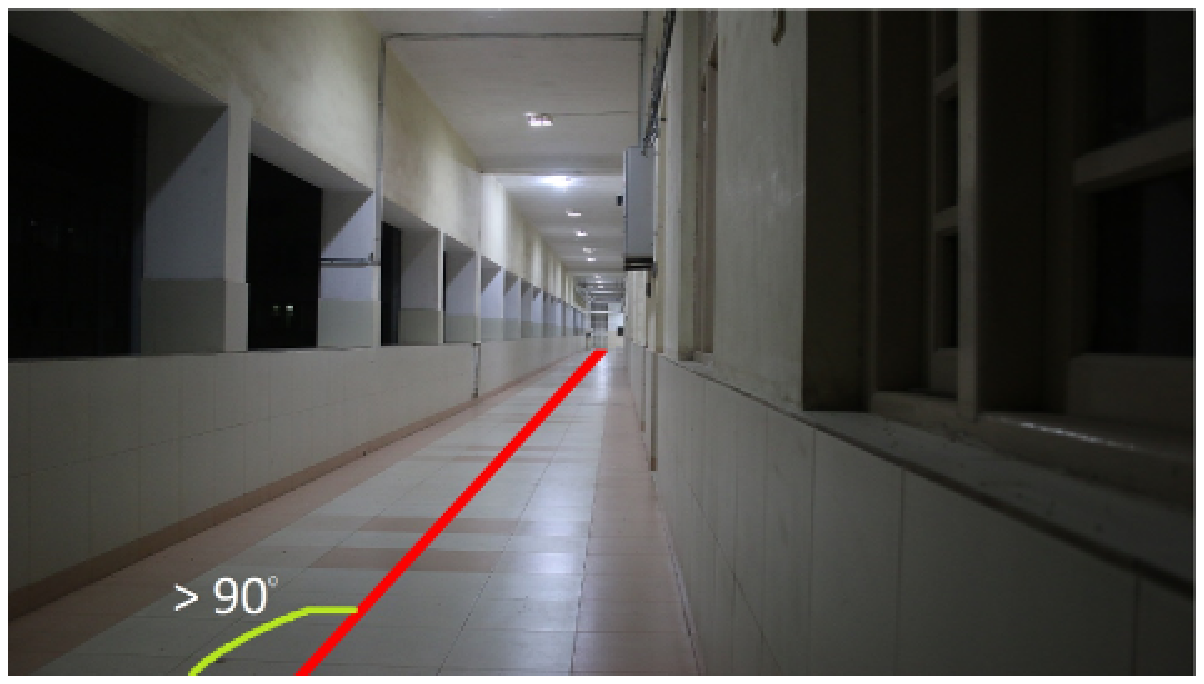}}
	\subfigure[On the CBL]{\includegraphics[width=.33\textwidth]{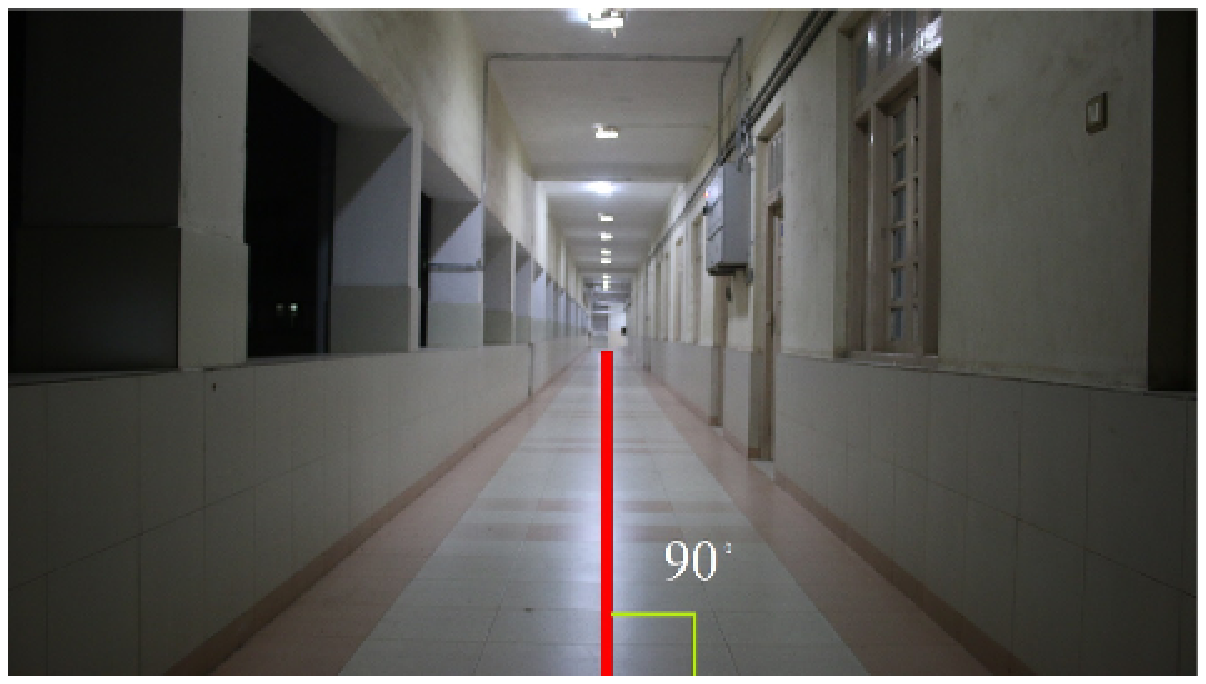}}
	\caption{Three different positions of the UAV over a horizontal line perpendicular to the CBL}
	\label{angle}
\end{figure*}

The bisector images are processed to detect the markers, and then these markers are connected by a red colored line, which forms the CBL on the image plane (Figure~\ref{dataset}). The bisector images are then checked manually for any discrepancy (CBL is not drawn properly), and erroneous images are discarded from the dataset. It has been observed that when the UAV is on the left side of the CBL, this red line on the image plane forms an acute angle with the bottom image boundary, as shown in Figure~\ref{angle}(a). In case the UAV is on the right side of the CBL, the angle formed is obtuse (Figure~\ref{angle}(b)). When the UAV is at the center, it forms a right angle (Figure~\ref{angle}(c)). Hence, for the corresponding actual image in the dataset, these angles (in radians), are stored as the target values. It may be noted that these angles remain the same irrespective of the UAV tilt. The angles lie in the range $[0,3.14]$. For safe localization of the UAV, always the goal is to keep the UAV at the center of the corridor. Alternatively, when the UAV is either on the left or right side of the CBL, our aim is to generate commands for the UAV such that this angle will shift towards the right angle. This labeled data will help in knowing the translational shift of the UAV from the CBL. However, it does not provide any information regarding the UAV tilt/orientation.  
\begin{figure*}[!htbp]   
	\subfigure[Aligned with the CBL]{\includegraphics[width=.33\textwidth]{center}}	
	\subfigure[Left tilted]{\includegraphics[width=.33\textwidth]{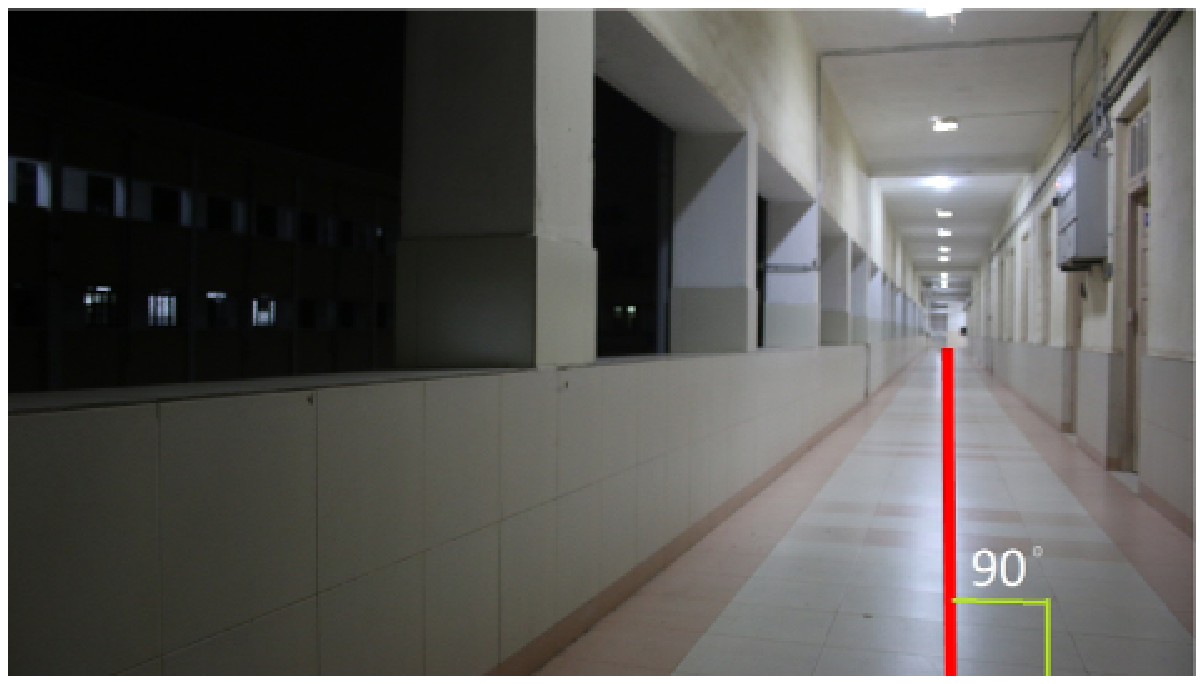}}
	\subfigure[Right tilted]{\includegraphics[width=.33\textwidth]{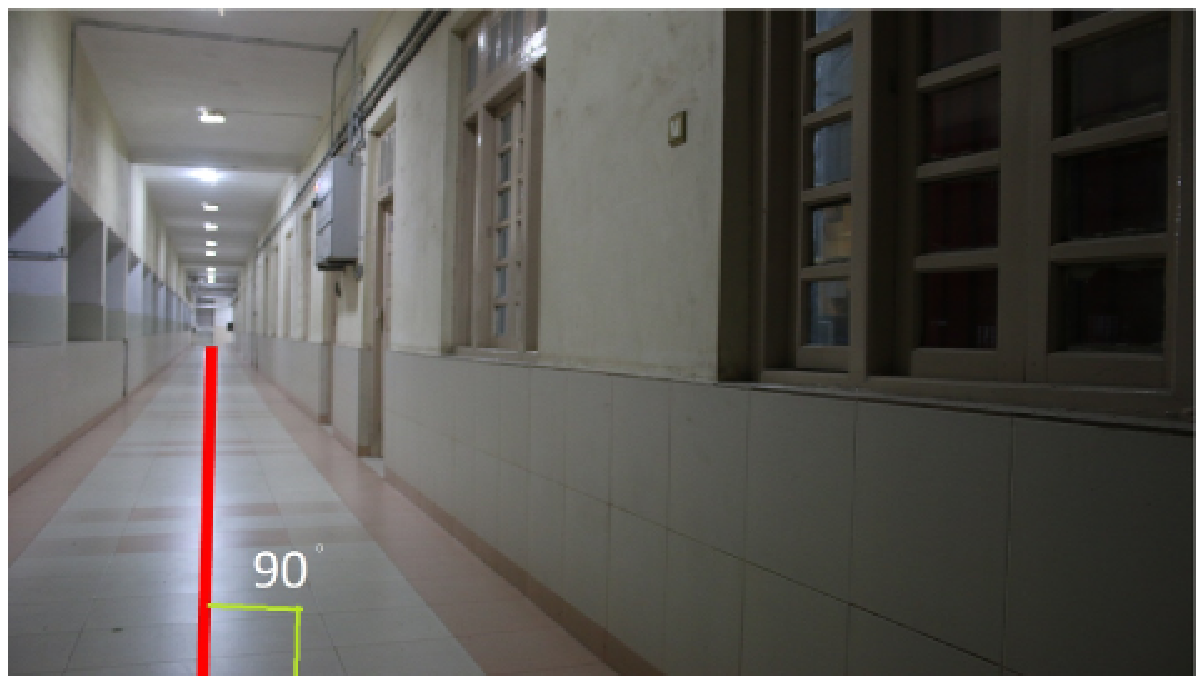}}
	\caption{Three different orientations of the UAV, when it is situated on the CBL}
	\label{distance}
\end{figure*} 
\par Once the UAV is at the center of the corridor over the CBL, we will try to measure the orientation of the UAV. It has been observed that when the UAV is at the center and aligned with the CBL, the red line on the bisector image cuts the image plane vertically into two equal halves, as shown in Figure~\ref{distance}(a). However, when the UAV is at the center but left tilted, this red line although perpendicular to the bottom image boundary deviates from the center and cuts the image plane into two unequal halves with the left side having more area than the right side (Figure~\ref{distance}(b)). The opposite scenario holds true, when the UAV is at the center, but right tilted (Figure~\ref{distance}(c)). We calculate the number of pixels in the x-direction from the left image boundary to the vertical red line and normalize it with the image width to obtain the target value for the corresponding actual image in the dataset. The value lies in the range of $[0, 1]$. This normalized value is known as the pixel distance of the CBL. When the distance is $0.5$, the UAV is properly aligned with the CBL. The UAV is right tilted when the distance is less than $0.5$, and vice versa. Hence, for the corresponding actual image in the dataset, these distances, are stored as the target values for measuring the orientation. Always, the goal is to keep the UAV aligned with the CBL. Alternatively, when the UAV is at the center, and either left or right tilted, our aim is to generate commands for the UAV such that this distance will shift towards $0.5$. This labeled data will help in knowing the rotational deviation of the UAV.
\par The dataset now contains two target values; (1) angle of the CBL (for rectifying the translational deviation), and (2) distance of the CBL (for rectifying the rotational deviation). Both target values can be utilized simultaneously in two similar DNNs to localize the UAV in corridor environments. We have shared this dataset as a public dataset, named NITRCorrV1~\cite{padhy2019dataset}.This dataset contains $35000$ training and $600$ testing images for angle, and 21000 training and 300 testing images for distance, and their corresponding target values.    
\section{Deep Neural Network for Autonomous UAV Navigation}\label{sec-dnn}
Our goal is to keep the UAV aligned with the CBL of the corridor throughout the process of navigation. This is a two-step process occurring consecutively; (1) Rectifying the translational deviation (side-wise variation) for bringing the UAV to center over the CBL, (2) Rectifying the rotational deviation (change in orientation) to align the UAV with the CBL, when the UAV is already at the center. Both processes are achieved by processing the images from the UAV front camera through pre-trained DNNs, that predict the deviations. Although the architecture of both the networks remains same, the target output differs; one is for angle in radian and other is for normalized pixel distance. As the unit of both the target values (angle and pixel distance) are different, it is not feasible to train both the quantities simultaneously using the same network.
\begin{table*}[!htbp]
	\centering
	\scriptsize
	\caption{Various pre-trained models and their augmented convolution layers to train our custom dataset. In the 2D convolution operation, Conv2d($C_{in}$,~$C_{out}$,~$k \times k$), $C_{in}$, $C_{out}$, and $k$ represent the number of input channels, number of output channels, and kernel size respectively.}
	\begin{tabular}{|c|c|c|c|c|}
		\hline
		Input & \begin{tabular}[c]{@{}c@{}}Pre-trained \\ Models\end{tabular} & \begin{tabular}[c]{@{}c@{}}Augmented \\ Convolution Layers\end{tabular} & \begin{tabular}[c]{@{}c@{}}Augmented last Fully \\ connected layer\end{tabular} & Output layer \\ \hline \hline
		\multirow{8}{*}{$320 \times 180$ } & AlexNet~\cite{krizhevsky2012imagenet} & \begin{tabular}[c]{@{}c@{}}no augmentation\\ of convolution layer\end{tabular} & $4096 \times 1$ & \multirow{8}{*}{$1 \times 1$} \\ \cline{2-4}
		& VGG-16~\cite{simonyan2014very} & \begin{tabular}[c]{@{}c@{}}Conv2d(512, 1024, $1 \times 1$)\\ Conv2d(1024, 128, $5 \times 5$)\\ Conv2d(128, 16, $1 \times 1$)\end{tabular} & $96 \times 1$ &  \\ \cline{2-4}
		& InceptionV3~\cite{szegedy2015going} & \begin{tabular}[c]{@{}c@{}}\textbf{Main: }Conv2d(2048, 1024, $1 \times 1$)\\ Conv2d(1024, 512, $2 \times 2$)\\ Conv2d(512, 128, $3 \times 3$)\\ \textbf{Aux: }Conv2d(768, 128, $4 \times 4$)\\ Conv2d(128, 32, $2 \times 2$)\end{tabular} & \begin{tabular}[c]{@{}c@{}}\textbf{Main: }$256 \times 1$\\ \textbf{Aux :}$640 \times 1$\end{tabular} &  \\ \cline{2-4}
		& ResNet-50~\cite{he2016deep} & \multirow{3}{*}{\begin{tabular}[c]{@{}c@{}}Conv2d(2048, 1024, $1 \times 1$)\\ Conv2d(1024, 128, $5 \times 5$)\\ Conv2d(128, 8, $1 \times 1$)\end{tabular}} & \multirow{3}{*}{$96 \times 1$} &  \\ \cline{2-2}
		& ResNet-101~\cite{he2016deep} &  &  &  \\ \cline{2-2}
		& ResNet-152~\cite{he2016deep} &  &  &  \\ \cline{2-4}
		& DenseNet-201~\cite{huang2016densely} & \begin{tabular}[c]{@{}c@{}}Conv2d(1920, 1024, $1 \times 1$)\\ Conv2d(1024, 128, $5 \times 5$)\\ Conv2d(128, 16, $1 \times 1$)\end{tabular} & \multirow{2}{*}{$96 \times 1$} &  \\ \cline{2-3}
		& DenseNet-161~\cite{huang2016densely} & \begin{tabular}[c]{@{}c@{}}Conv2d(2208, 1024, $1 \times 1$)\\ Conv2d(1024, 128, $5 \times 5$)\\ Conv2d(128, 16, $1 \times 1$)\end{tabular} &  &  \\ \hline
	\end{tabular}
	\label{Deep-models}
\end{table*}    
\subsection{Network Structure}
For both the tasks, we used some of the standard pre-trained models (AlexNet~\cite{krizhevsky2012imagenet}, VGGNet~\cite{simonyan2014very}, InceptionNet~\cite{szegedy2015going}, ResNet~\cite{he2016deep}, DenseNet~\cite{huang2016densely}) to train our custom dataset. The last fully connected layer of the pre-trained model is replaced by a few convolution layers to decrease the output feature map size. At the end of the network, one fully connected layer is appended to obtain $1 \times 1$ output. The input image resolution is $320 \times 180$. Various networks that are used to train our dataset is shown in Table~\ref{Deep-models}. The proposed network acts as a regression model, as the network directly predicts a numeric value, instead of a class. The same network is trained separately for translational deviation and rotational deviation, just by changing the target output. In case of translational deviation, the target is the angle (in radians) made by the CBL with bottom image boundary plane, and for rotational deviation, the target is a normalized pixel distance of the CBL from the left image boundary. The angle lies in the range $[0, 3.14]$ and the normalized pixel distance lies in the range of [0, 1]. A free body diagram of our proposed model is shown in Figure~\ref{free-body}.
\subsection{Loss Function}
Accurate selection of a loss function plays an important role in carrying out any deep learning based optimization task. We have used the Mean Absolute Error (MAE) loss function to train our models. MAE is formulated as ---
\begin{equation}\label{eq:loss}
\mathrm{MAE}(\hat{y} ,y) = \frac{1}{n}\sum_{i=1}^{n}|\hat{y_i} -y_i|
\end{equation}
Where, $\hat{y}$ and $y$ denote the list of predicted and target values for a mini batch of size $n$.
\subsection{Training}
\noindent\textbf{Input Pre-processing.} The proposed model accepts an input of resolution $320 \times 180$. Before processing, the original BGR pixel values, which lie in the range $[0, 255]$, are first normalized to the range $[0, 1]$. Each of the Blue, Green and Red channels are then normalized by the corresponding mean ($\mu_b = 0.406, \mu_g = 0.456, \mu_r = 0.485$) and standard deviation ($\sigma_b = 0.225, \sigma_g = 0.224, \sigma_r = 0.229$) of the ImageNet classification dataset~\cite{deng2009imagenet}. Prediction of translational deviation requires images from all possible locations of the corridor, however rotational deviation requires images when the UAV is at the center and tilted in all three orientations (left, right or center).\\
\noindent\textbf{Parameter Initialization.} The initial parameters of the networks remain same for both the tasks. MAE~\ref{eq:loss} is used as the loss function, while Stochastic Gradient Descent (SGD)~\cite{bottou2010large} is used as the optimizer. Overfitting is handled by $L2$ regularization. Initial Momentum is set to $0.9$ and learning rate starts at $0.001$. 
\par Training is performed separately for the prediction of translational and rotational deviation. We used 9 different pre-trained networks (AlexNet, VGG-16, InceptionV3, ResNet-50, ResNet-101, ResNet-152, DenseNet-201, DenseNet-161) to train our model. Training is performed separately for each of the pre-trained networks. Learning rate is reduced by a factor of $5$, if the training loss remained same for consecutive $5$ iterations. The process of training continues, till the learning rate reaches $10^{-15}$. 
\subsection{Control Command generation}
When the UAV is properly aligned with the CBL of the corridor, there is minimal chance of collision. Depending on the predictions of both the networks, control commands are generated to localize the UAV along the CBL of the corridor. The images from the UAV front camera are continuously processed through the trained networks to maintain the UAV pose. Algorithm~\ref{algo:1} depicts the flow for generating the control commands for UAV navigation through an indoor corridor.
\begin{algorithm}[!ht]
	\caption{Control command generation}
	\label{algo:1}
	\SetKwInput{KwInput}{Input}
	\SetKwInput{KwOutput}{Output}
	\SetKwInput{KwInit}{Initialisation}
	\KwInput{Image From UAV front camera: $img$}
	\KwOutput{UAV direction: $[pitch, roll, yaw]$}
	$angle$ = TrainedModelForAngle($img$)\\
	\uIf {$angle~out~of~bound~for~continously~1~second$} 
	{
		Land the UAV
	}
	\uIf {$angle \approx 90 \degree$} 
	{
		dist = TrainedModelForDistance($img$)\\
		\uIf {$dist \approx 0.5$}
		{
			Actuate UAV in Pitch Forward
		}
		\uElseIf {$dist < 0.5 - \delta$}
		{
			Actuate UAV in Yaw Left until $dist \approx 0.5$
		}
		\Else
		{
			Actuate UAV in Yaw Right until $dist \approx 0.5$
		}
	}
	\uElseIf {$angle < 90 \degree - \delta$}
	{
		Actuate UAV in Roll Right until $angle \approx 90 \degree$
	}
	\Else
	{
		Actuate UAV in Roll Left until $angle \approx 90 \degree$
	}
	\KwRet{$[pitch, roll, yaw]$}
\end{algorithm}	 

\section{Experiments}\label{sec-expt}
This section presents a detailed analysis of all the experiments carried out for the task of UAV navigation inside indoor corridor environments, using DNNs. The test sets of our custom dataset, NITRCorrV1, is used for validation purpose. We also show the system setup and hardware platform on which the experiments are being carried out. Finally, we show the results of our proposed navigation algorithm in real-world UAV flight experiments.
\subsection{System Setup}
We used PyTorch deep learning library to implement our deep networks. Training is performed with GTX 1080 Ti graphics card, which has a memory of $11$GB. ILSVRC~\cite{russakovsky2015imagenet} pretrained weights are used for initializing the standard state-of-the-art networks. Weights for other convolution layers are initialized from a normal distribution $\mathcal{N}(0,\,\sqrt{2/ck^2})$, where $k$ represents the filter size and $c$ is the number of output channels of the convolution operation.
\subsection{Evaluation Metrics}
Evaluation is done on the NITRCorrV1 labeled test set, which contains 600 images for translational and 300 images for rotational deviation, respectively. Test input images are preprocessed using the same procedure as the training images. We used three different evaluation metrics to validate our proposed DNN models.\\
\textbf{(a)} $\mathrm{Mean~Squared~Error:~MSE}(\hat{y} ,y) = \frac{1}{n}\sum_{i=1}^{n}(\hat{y_i} -y_i)^2$\\
\textbf{(b)} $\mathrm{Mean~Absolute~Error:~MAE}(\hat{y} ,y) = \frac{1}{n}\sum_{i=1}^{n}|\hat{y_i} -y_i|$\\
\textbf{(c)} $\mathrm{Mean~Relative~Error:~MRE}(\hat{y} ,y) = \frac{1}{n}\sum_{i=1}^{n}\frac{|\hat{y_i} -y_i|}{y_i}$\\
Where, $\hat{y}$ and $y$ denote the list of predicted and target values for a mini batch of size $n$.
\subsection{Results and Discussion}
\begin{figure*}[!htbp]
	\centering
	\caption{Qualitative performance evaluation of translational deviation for different corridor locations of National Institute of Technology, Rourkela, India. Ground truth and predicted values are given in degree. GT: Ground truth, PR: Predicted.}
	\begin{tabular}{cccc}
		\textbf{\begin{tabular}[c]{@{}c@{}}TIIR Building\end{tabular}} & \textbf{\begin{tabular}[c]{@{}c@{}}Physics\\ Department\end{tabular}} & \textbf{\begin{tabular}[c]{@{}c@{}}Life Science\\ Department\end{tabular}} & \textbf{\begin{tabular}[c]{@{}c@{}}Computer Science\\ Department\end{tabular}} \\  		
		\includegraphics[width=0.23\textwidth]{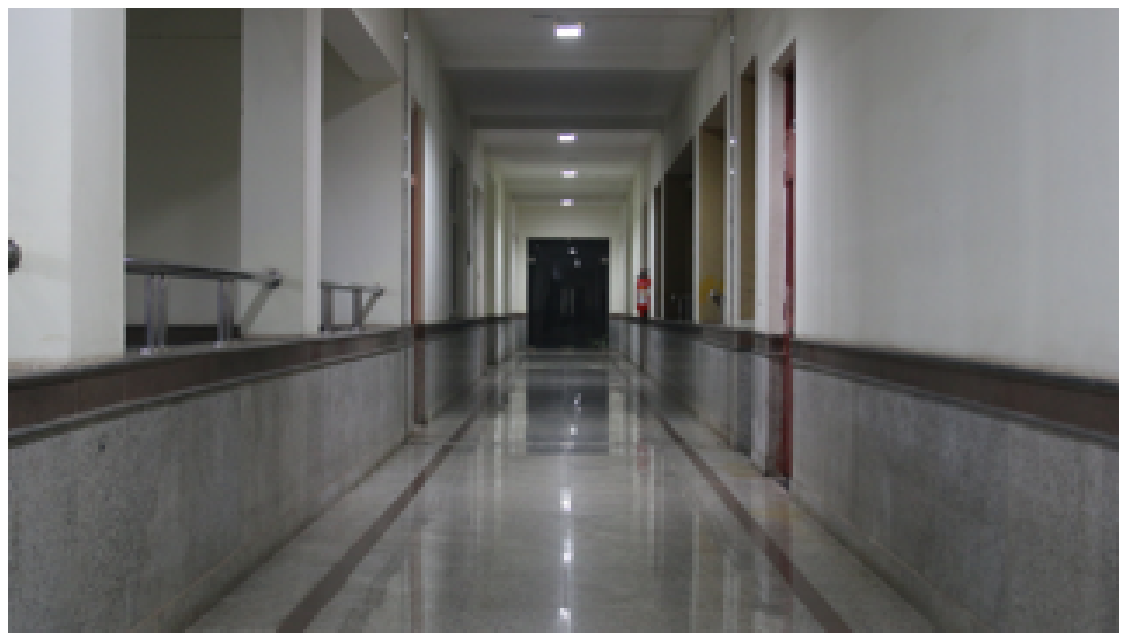} & \includegraphics[width=0.23\textwidth]{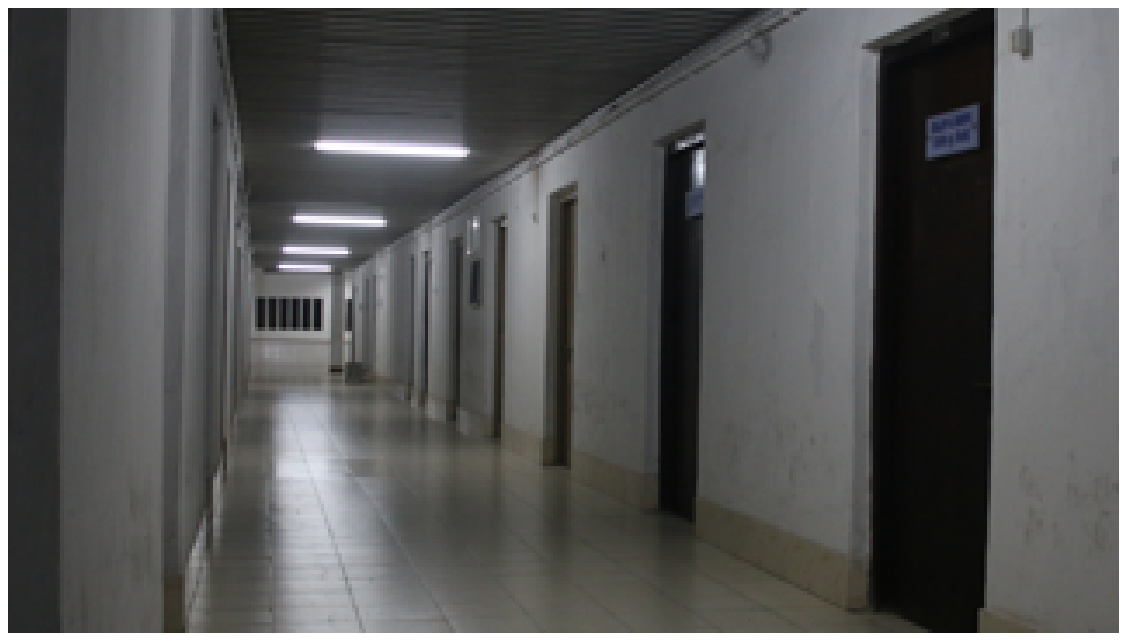} & \includegraphics[width=0.23\textwidth]{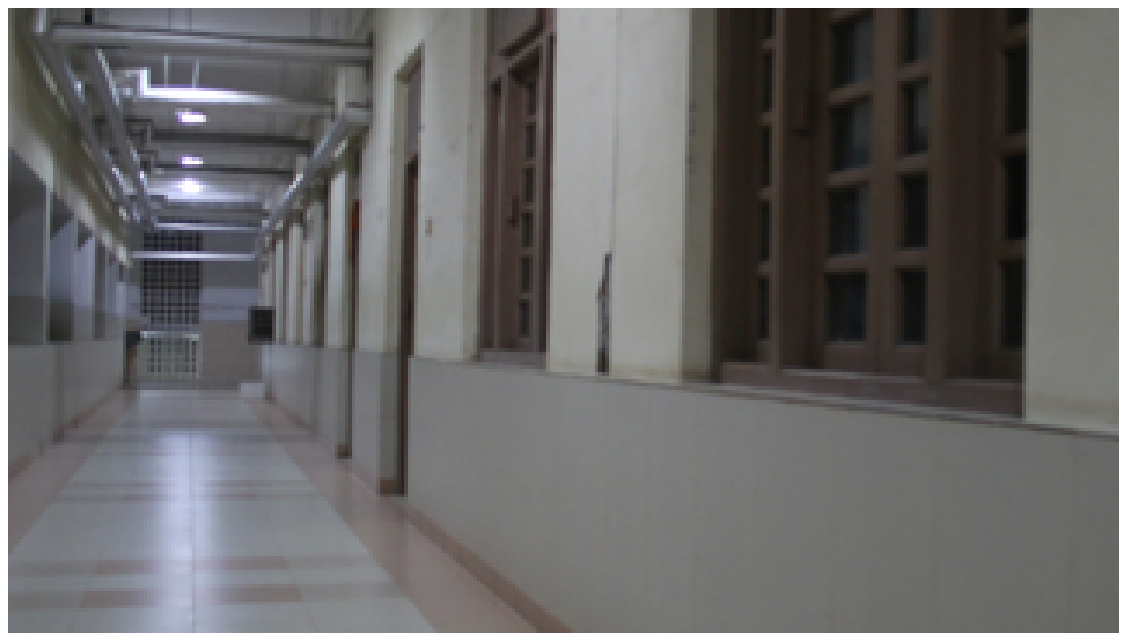} & \includegraphics[width=0.23\textwidth]{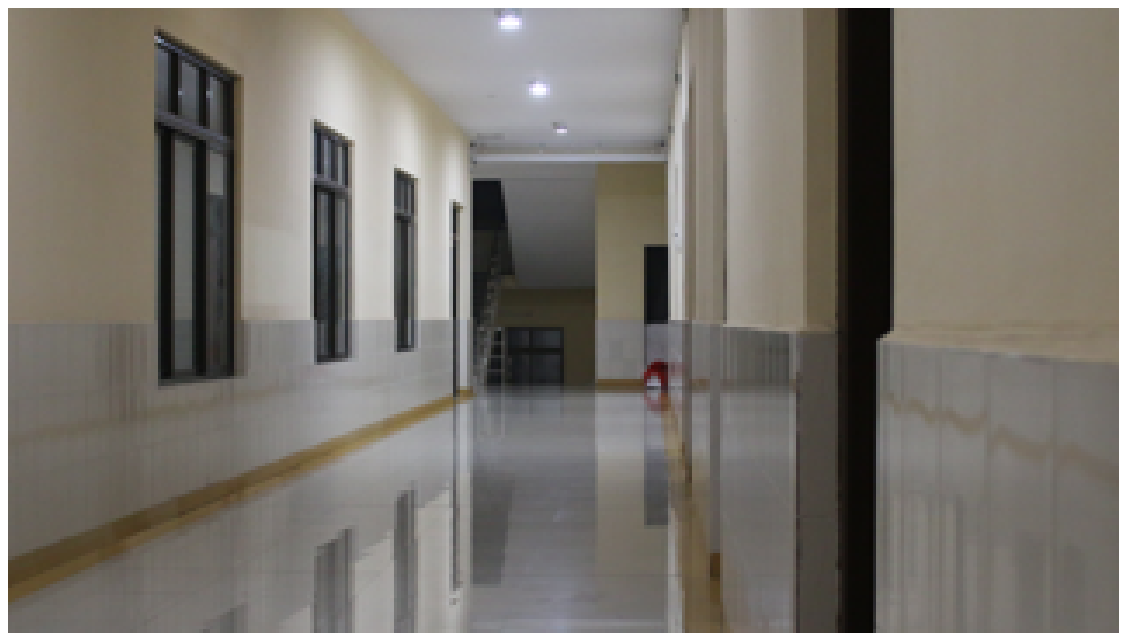} \\
		\textbf{\begin{tabular}[c]{@{}c@{}}GT: 90.720\degree\\ PR: 90.303\degree\end{tabular}} & \textbf{\begin{tabular}[c]{@{}c@{}}GT: 47.055\degree\\ PR: 46.707\degree\end{tabular}} & \textbf{\begin{tabular}[c]{@{}c@{}}GT: 90.000\degree\\ PR: 91.142\degree\end{tabular}} & \textbf{\begin{tabular}[c]{@{}c@{}}GT: 136.507\degree\\ PR: 135.253\degree\end{tabular}}\\ \\ \hline
	\end{tabular}
	\label{trans-qual}
\end{figure*}

\begin{figure*}[!htbp]
	\centering
	\caption{Qualitative performance evaluation of rotational deviation for different corridor locations of National Institute of Technology, Rourkela, India. Ground truth and predicted values are in the range $[0, 1]$. GT: Ground truth, PR: Predicted.}
	\begin{tabular}{cccc}
		\textbf{\begin{tabular}[c]{@{}c@{}}TIIR Building\end{tabular}} & \textbf{\begin{tabular}[c]{@{}c@{}}Physics\\ Department\end{tabular}} & \textbf{\begin{tabular}[c]{@{}c@{}}Life Science\\ Department\end{tabular}} & \textbf{\begin{tabular}[c]{@{}c@{}}Computer Science\\ Department\end{tabular}} \\    
		\includegraphics[width=0.23\textwidth]{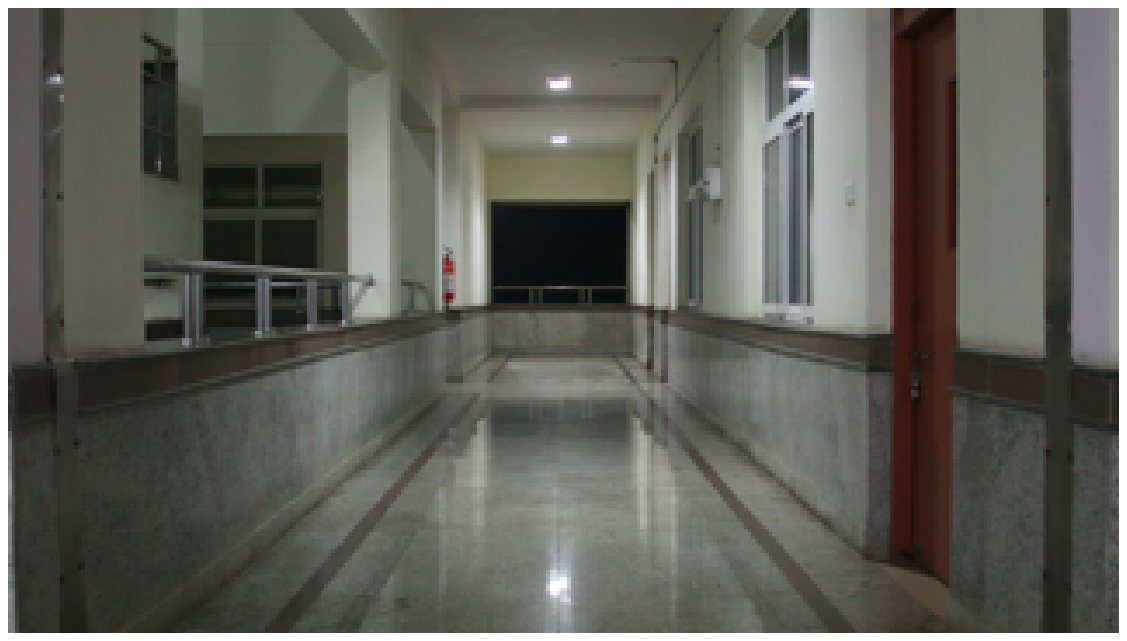} & \includegraphics[width=0.23\textwidth]{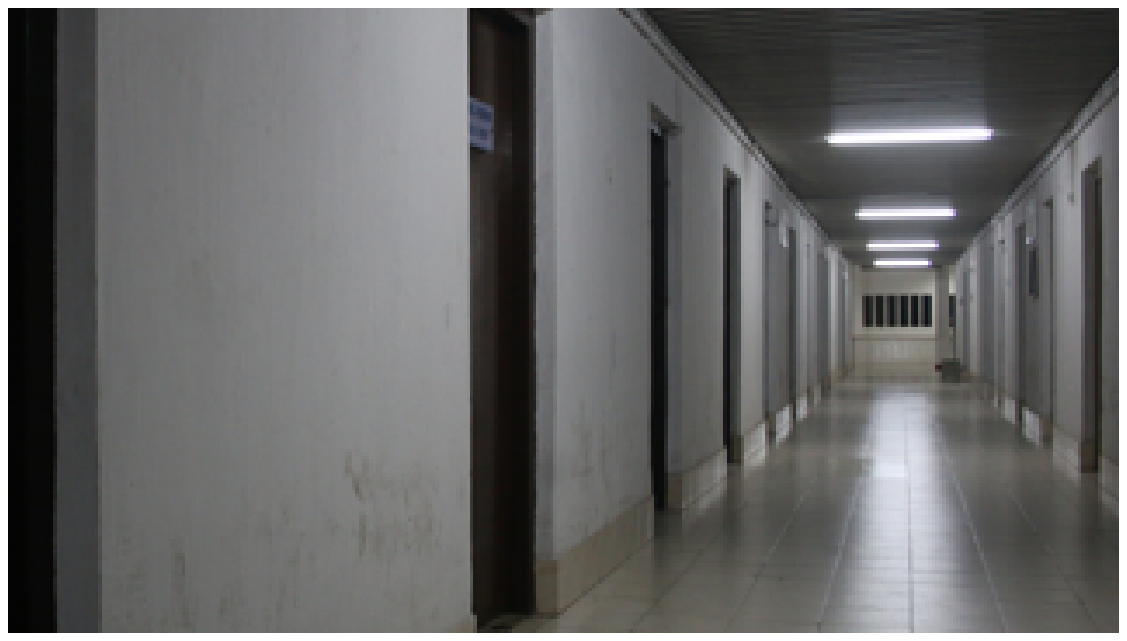} & \includegraphics[width=0.23\textwidth]{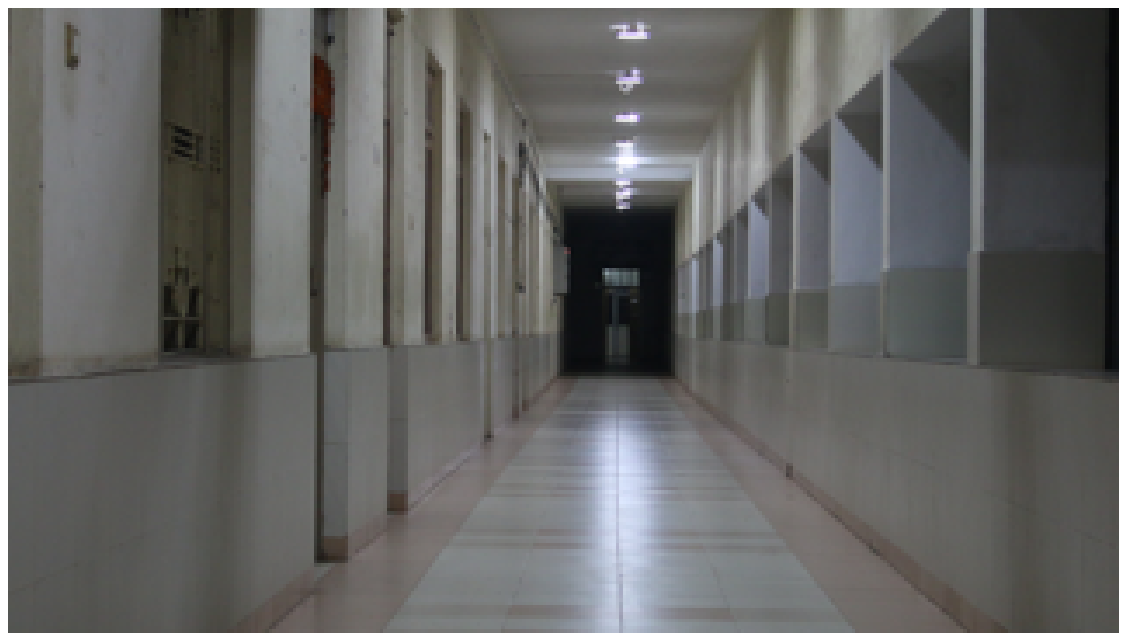} & \includegraphics[width=0.23\textwidth]{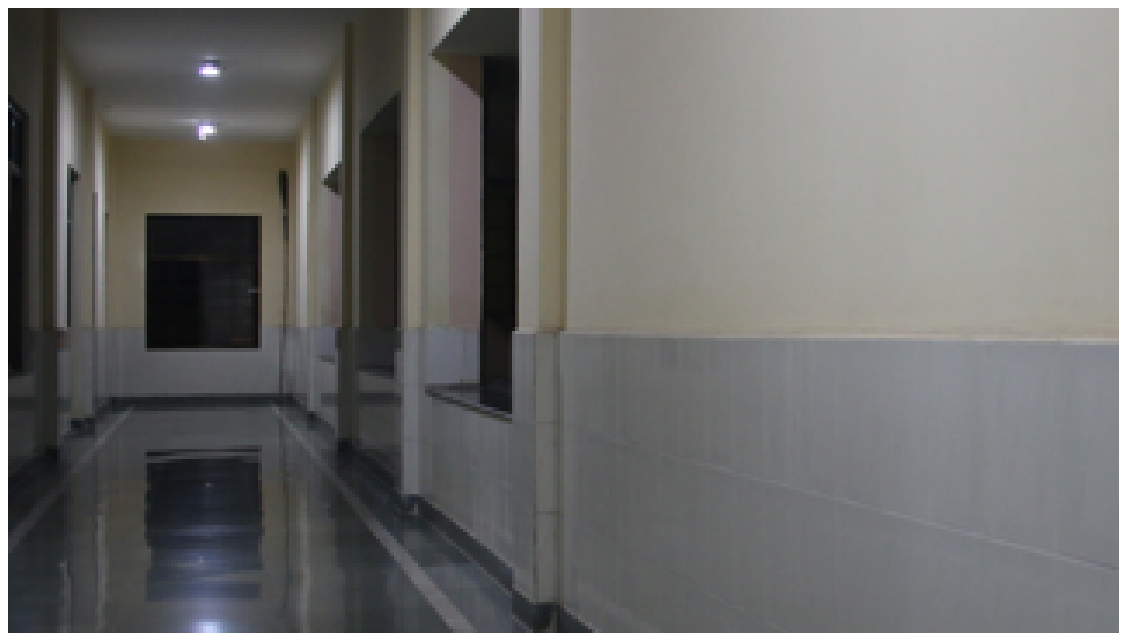} \\
		\textbf{\begin{tabular}[c]{@{}c@{}}GT: 0.488\\ PR: 0.491\end{tabular}} & \textbf{\begin{tabular}[c]{@{}c@{}}GT: 0.805\\ PR: 0.808\end{tabular}} & \textbf{\begin{tabular}[c]{@{}c@{}}GT: 0.547\\ PR: 0.528\end{tabular}} & \textbf{\begin{tabular}[c]{@{}c@{}}GT: 0.167\\ PR: 0.163\end{tabular}} \\ \\ 
	\end{tabular}
	\label{rotate-qual}
\end{figure*}

We compared the results of our proposed model using $9$ different augmented pretrained networks, as given in Table~\ref{Deep-models}. The comparison is done separately for translational and rotational deviation using separate test sets. Table~\ref{trans} and Table~\ref{rotate} show the quantitative comparison results for both the tasks. It can be viewed that DenseNet-161 performs superior as compared to other pretrained networks for both the tasks in terms of all the evaluation metrics. Hence, for real-time navigation experiments, we used DenseNet-161 pretrained network to navigate the UAV in corridor environments. Location wise qualitative comparison results using DenseNet-161 pretrained network are shown in Tables~\ref{trans-qual} and \ref{rotate-qual}.
\begin{table}[!htbp]
	\centering
	\scriptsize
	\caption{Evaluation metrics for the prediction of translational deviation of the UAV using 9 different pretrained networks over $600$ test images.}
	\begin{tabular}{|c|c|c|c|}
		\hline
		\begin{tabular}[c]{@{}c@{}}Pre-trained\\ Model\end{tabular} & MSE & MAE & MRE \\ \hline \hline
		AlexNet & 0.21997 & 1.72831 & 1.39280 \\ 
		VGG-16 & 0.47318 & 2.84597 & 2.41795 \\ 
		InceptionV3 & 0.11929 & 1.44906 & 1.20959 \\ 
		ResNet-50 &  &  &  \\ 
		ResNet-101 & 0.11103 & 1.46875 & 1.17946 \\ 
		ResNet-152 & 0.11032 & 1.41558 & 1.16529 \\ 
		DenseNet-201 & 0.12383 & 1.79144 & 1.42709 \\ 
		DenseNet-161 & \textbf{0.05791} & \textbf{1.32693} & \textbf{1.08712} \\ \hline
	\end{tabular}
	\label{trans}
\end{table}
\begin{table}[!htbp]
	\centering
	\scriptsize
	\caption{Evaluation metrics for the prediction of rotational deviation of the UAV using 9 different pretrained networks over $300$ test images.}
	\begin{tabular}{|c|c|c|c|}
		\hline
		\begin{tabular}[c]{@{}c@{}}Pre-trained\\ Model\end{tabular} & MSE & MAE & MRE \\ \hline \hline
		AlexNet & 5.5677 & 27.0064 & 54.1467 \\ 
		VGG-16 &  &  &  \\ 
		InceptionV3 & 0.0687 & 3.1364 & 10.4852 \\ 
		ResNet-50 & 0.0473 & 2.7485 & 9.0729 \\ 
		ResNet-101 & 0.1186 & 4.2163 & 14.6806\\ 
		ResNet-152 & 0.06617 & 3.4258 & 10.8230 \\ 
		DenseNet-201 & 0.0828 & 3.6442 & 12.1421 \\ 
		DenseNet-161 & \textbf{0.0326} & \textbf{2.5060} & \textbf{1.557} \\ \hline
	\end{tabular}
	\label{rotate}
\end{table}

\subsection{Real World Navigation Experiments}
We tested our deep learning based UAV navigation algorithm in real-world indoor corridors. Parrot A.R.Drone quadcopter is used to validate the experiments. It has a static front camera, which provides video feed at 30 frames per second (fps). The UAV is capable of producing a fisheye video ($92\degree$ field of view) with a resolution of $640 \times 480$. Before processing the images through the DNN, the fisheye images are converted to their rectilinear equivalent. Also, the images are downsampled to $320 \times 180$, which is the default input to the proposed deep learning model. The UAV is connected to a ground-based system, where the deep learning model is running, though wireless LAN and communication take place using the Robot Operating System (ROS). Image frames from the UAV front camera are passed continuously through the ROS to the deep learning system. Image transmission delay through ROS is about $\mathrm{0.21s}$. Our deep learning model prediction takes about $\mathrm{0.08s}$ for measuring both translational and rotational deviation simultaneously. The system then generates the appropriate control command using Algorithm~\ref{algo:1}, and forwards it to the UAV through the ROS. The UAV internal PID controller is then responsible for executing the command and maintaining safe UAV pose during its navigation. It may be noted that apart from our network prediction, different factors, such as control and state estimation affect the actual UAV flight in real-world scenarios. Because of the unstable and fluctuating state estimation of Parrot drone, we cannot employ the positional feedback in path planning, which sometimes deteriorates the whole navigation process. 
It is observed that our proposed DNN based navigation algorithm works with good accuracy in most of the corridors.    	
\section{Conclusion}\label{sec-conclusion}
In this work, we proposed a DNN based UAV localization algorithm, which enables the UAV to navigate safely inside corridors environments. Unlike the previous methods, where the deep models were designed to predict flight commands directly, our proposed method makes use of an important characteristic of a corridor, the CBL, to generate commands. We designed a single deep network and trained it separately for two tasks; (a) prediction of translational deviation of the UAV from the CBL, and (b) prediction of rotational deviation of the UAV with respect to the CBL. These two predictions are then used to design a navigation algorithm, which enables the UAV to maintain a safe pose inside the corridor. We also provide a new corridor dataset, NITRCorrV1, in which the corridor images are labeled in terms of translational and rotational deviation values. Our proposed navigation algorithm is capable of navigating a UAV inside real-world corridor environments with good accuracy.  	

\bibliographystyle{IEEEtran}
\bibliography{obstacle}
\end{document}